\documentclass[runningheads]{llncs}

\usepackage[T1]{fontenc}
\usepackage{graphicx}
\usepackage{mathtools}
\usepackage{float}
\usepackage{amsfonts}

\begin{document}
\title{FPED: A Functional-Network Prior-Guided Mixture-of-Experts Framework for Interpretable Brain Decoding}
\titlerunning{FPED: Brain Decoding with MoE and Functional Networks}
%
\author{Yudan Ren\and
Pengcheng Shi  \and Zihan Ma \and Xiaowei He \and Xiao Li 
}
\authorrunning{Ren et al.}
%
\institute{School of Electronic Information (School of Artificial Intelligence), Northwest University, Xi'an 710127, China\\
\email{\{yudan.ren@nwu.edu.cn,spc@stumail.nwu.edu.cn,202233494@stumail.nwu.edu.cn,\\hexw@nwu.edu.cn,lixiao@nwu.edu.cn\}}
}
\maketitle              

\begin{abstract}
Visual image reconstruction from functional Magnetic Resonance Imaging (fMRI) 
is a fundamental task in brain decoding, providing a crucial pathway for 
understanding human perceptual mechanisms and developing advanced brain-computer 
interfaces (BCIs). However, most current methods simply flatten fMRI signals 
from localized visual cortices into one-dimensional (1D) vectors, mapping them 
directly into latent spaces such as that of Contrastive Language-Image 
Pre-training (CLIP). This paradigm not only disrupts the inherent network 
topology of the brain—leading to limited neuroscientific interpretability—but 
also overlooks the synergistic contributions of other distributed functional 
networks in processing high-level visual semantics.

To address these limitations, we propose FPED, a Functional-Network Prior-Guided Mixture of Experts (MoE) framework for interpretable brain decoding. FPED explicitly models different functional brain networks as specialized experts and employs adaptive routing to capture their complementary contributions to visual semantic understanding. Unlike conventional homogeneous decoding paradigms, our framework incorporates neurobiologically grounded priors to enable structured and interpretable network-level representation learning. Experimental results demonstrate that FPED achieves highly competitive semantic reconstruction performance with only 0.68B parameters. The learned routing dynamics reveal biologically meaningful correspondence between functional brain networks and modality-specific semantic processing, providing transparent neuroscientific interpretability. This suggests that brain network-aware expert modeling is a promising direction for bridging neural decoding and biologically inspired artificial intelligence.

\end{abstract}

\section{Introduction}
Functional Magnetic Resonance Imaging (fMRI) is a non-invasive functional brain imaging method that indirectly measures neuronal activity by monitoring Blood-Oxygen-Level-Dependent (BOLD) signals \cite{ref_Liu2025}. The intrinsic connection between neural decoding and human visual understanding mechanisms makes fMRI-based visual reconstruction a pivotal frontier at the intersection of cognitive neuroscience and computer vision \cite{ref_Du2022,ref_ZhaoL2023}. From a neural decoding perspective, this task aims to reconstruct external visual stimuli from brain activity, revealing the encoding principles and representational forms of visual information within the brain \cite{ref_Naselaris2011}. From a neuroscientific perspective, visual understanding is a hierarchical and network-distributed process that progresses from the perception of low-level features to the processing of high-level semantic content \cite{ref_Zafar2015}. However, existing studies have largely focused on improving reconstruction fidelity from a purely engineering perspective, often overlooking the necessity for semantically consistent synthesis that aligns with biological brain mechanisms. Consequently, this focus has led to a significant lack of structural and functional interpretability, failing to elucidate how distributed neural networks collectively contribute to visual understanding.

With the development of diffusion models and multimodal contrastive learning methods such as Contrastive Language-Image Pre-training (CLIP), the field of visual decoding has achieved significant technological strides \cite{ref_Takagi2023,ref_Ferrante2023}. However, most current mainstream methods process fMRI signals by simply flattening data from visual-related brain regions into one-dimensional vectors and directly aligning them with CLIP feature spaces \cite{ref_Scotti2023,ref_Wang2024}. This localized processing approach suffers from two critical limitations: (1)  it disregards the large-scale brain network organization and its functional specialization, failing to account for how visual understanding arises from the synergistic contributions of multiple distributed functional networks; (2)  it overlooks the essential contributions of higher-order cognitive networks, such as the default mode network (DMN) and dorsal attention network (DAN), in visual semantic interpretation—neglecting these roles often leads to unstable or inaccurate semantic reconstruction, as the model lacks the necessary top-down constraints to prevent noise-driven semantic drifting.

To address these limitations, we propose employing a joint encoding strategy across multiple brain functional networks \cite{ref_Bonner2017}, modeling brain activity as a composition of heterogeneous yet complementary functional networks rather than a homogeneous feature vector. Given the significant heterogeneity across networks in processing characteristics, temporal dynamics, and functional specialization, simple feature concatenation or uniform fusion strategies are inadequate for adaptively leveraging network-specific representations. Therefore, inspired by the Mixture-of-Experts (MoE) paradigm \cite{ref_Masoudnia2014}, we treat each network as an expert module, employing a learnable gating mechanism to dynamically weight their contributions instead of relying on fixed-weight concatenation.

In this study, we propose FPED, a brain functional network prior-guided Mixture-of-Experts decoding framework. Unlike existing MoE-based brain decoding methods, FPED explicitly models large-scale functional brain networks as specialized experts and leverages adaptive routing to capture their complementary contributions to visual semantic understanding. The framework adopts a hierarchical expert architecture that first captures coarse-grained functional specialization across networks and then performs fine-grained integration through adaptive fusion, mimicking the hierarchical organization of human visual processing. Finally, the fused representation is aligned with a semantic latent space to drive high-fidelity visual reconstruction. Furthermore, a dynamic routing mechanism is introduced to model temporal variations in brain activity, enabling flexible adaptation to time-varying neural responses.

Our contributions are as follows:

\begin{itemize}
\item We propose FPED, a brain decoding framework that explicitly models the functional specialization and synergistic mechanisms of brain networks through a prior-guided Mixture-of-Experts architecture.

\item We evaluate FPED under both single-subject and multi-subject settings, validating its capacity to capture generalized representations across individuals while maintaining high decoding performance.

\item Through extensive comparative experiments, we demonstrate the necessity of network-level priors and the performance advantages of whole-brain multi-network modeling over traditional visual cortex-centric approaches.

\item We provide a comprehensive interpretability analysis that bridges decoding performance with neuroscientific mechanisms, elucidating how different brain networks contribute to visual semantic understanding.
\end{itemize}

\section{Related Works}

\subsection{fMRI to Image Reconstruction}
Advances in fMRI technology and large-scale datasets have fueled significant progress in fMRI-to-image reconstruction. While early generative models (VAEs, GANs) struggled with balancing semantic fidelity and spatial precision \cite{ref_Goodfellow2020,ref_Du2018,ref_Ren2021,ref_Lin2022}, recent approaches leveraging diffusion models and CLIP alignment have substantially improved reconstruction quality \cite{ref_Scotti2023,ref_Wang2024,ref_MindBridge2024,ref_MindEye2}.Some researchers integrate multi-level CLIP features, eliminating the need for independent VAE pathways\cite{ref_XiaT2026}. However, most methods oversimplify fMRI data by flattening signals from visual regions and aligning them directly with global semantic embeddings. This methodology overlooks the structured, distributed nature of brain activity, resulting in suboptimal modeling of inter-regional dependencies and limited semantic robustness. Furthermore, the predominant focus on visual cortices neglects the critical role of higher-order cognitive networks in shaping visual perception, restricting the capture of high-level conceptual information. The lack of explicit modeling of large-scale network interactions motivates the need for network-aware decoding frameworks.
\subsection{Multi-Functional Organization of the Visual Brain}
Visual cognition relies on coordinated interactions among multiple functional brain networks: the ventral visual stream for object recognition \cite{ref_Farivar2009}, the DAN for spatial attention \cite{ref_Machner2022}, and the DMN for semantic inference \cite{ref_Li2014}. Despite their functional specialization, these networks exhibit tight functional coupling during visual processing \cite{ref_DiCarlo2012}. However, contemporary fMRI-to-image reconstruction approaches seldom account for such multi-network collaborative dynamics, typically emphasizing visual cortex processing or adopting simplistic integration schemes. These limitations underscore the need for decoding frameworks that simultaneously capture network-level specialization and enable adaptive integration across distributed brain systems, motivating the adoption of expert-based modeling paradigms.
\begin{figure}
\includegraphics[width=\textwidth]{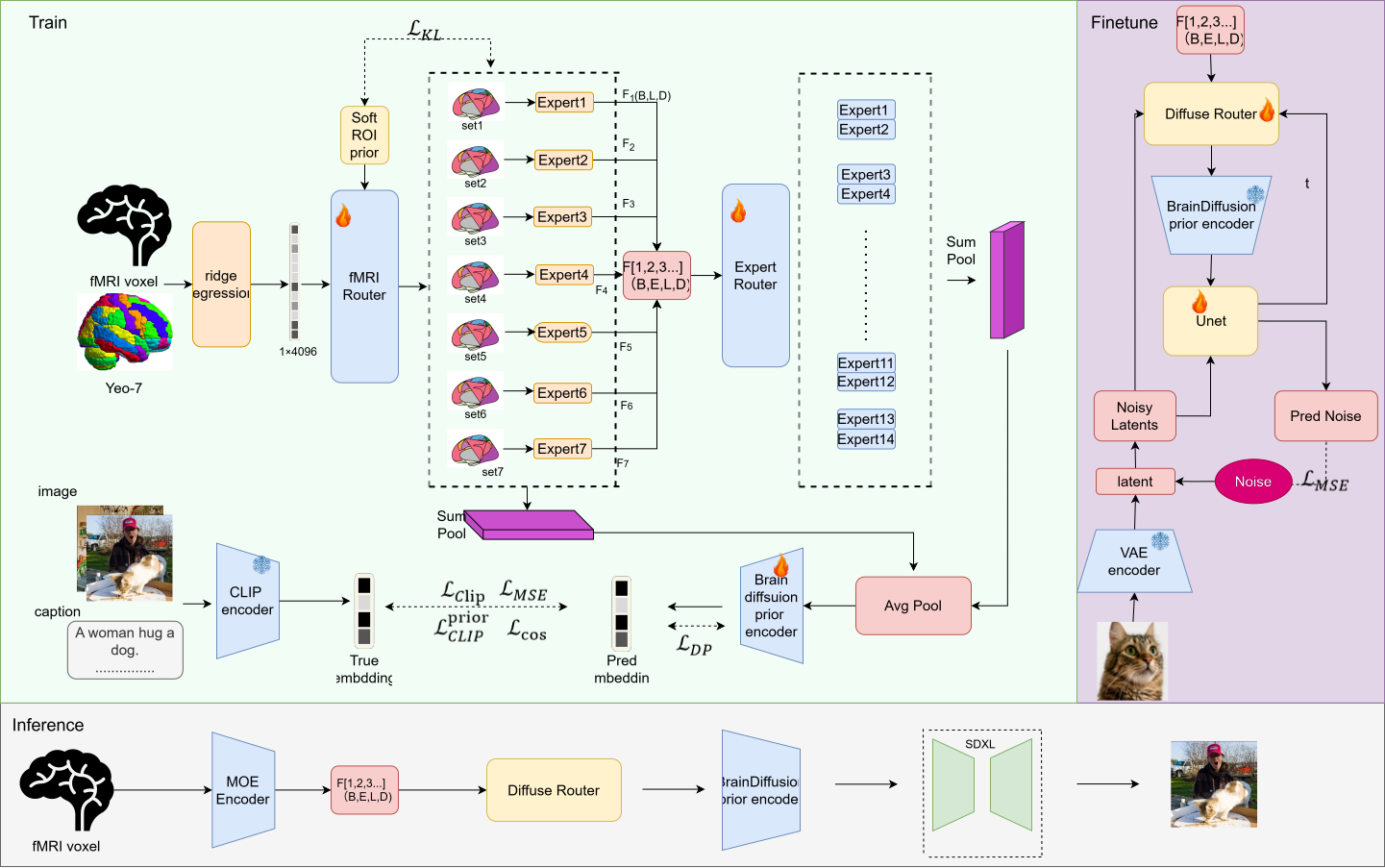}
\caption{Overview of the FPED framework. Stage 1 establishes a mapping 
from multi-network fMRI signals to CLIP embedding space. Stage 2 incorporates 
spatiotemporal routing mechanisms to enable dynamic gating of features across 
temporal brain activity variations.} 
\label{fig1}
\end{figure}
\section{Method}
Our proposed FPED framework comprises two sequential stages for progressive fMRI-to-image reconstruction, as illustrated in Fig 1.
\subsection{fMRI Feature Mapping}
\subsubsection{Data Preprocessing and Brain Registration}
We register whole-brain 3D fMRI data from the Natural Scenes Dataset (NSD)~\cite{ref_Allen2022} to the Yeo-7 functional network template~\cite{ref_YEOBT2011}. Using a per-sample top-k strategy~\cite{ref_Y2023}, we independently select, for each fMRI sample, the 14,877 voxels with the highest signal intensity; this top-k mask is computed and applied separately for every sample during training, validation, and testing. Within each functional network, ridge regression is applied voxel-wise for denoising; ridge coefficients are estimated on the training set and then frozen for validation and testing to avoid information leakage. The denoised voxel features from the seven functional networks are then combined (via the same network-wise aggregation used during training) and concatenated proportionally to form the $\mathbf{x}\in\mathbb{R}^{4096}$ brain feature vector as input to the routing mechanism.
\subsubsection{Hierarchical Mixture of Experts Architecture}

Our hierarchical routing architecture consists of two layers of expert mechanisms. The first layer comprises seven experts, each corresponding to one of the seven functional networks defined by the Yeo-7 atlas, while the second layer incorporates 14 experts designed for finer-grained feature processing. This progressive design enables the model to transition from coarse brain region representations to specific functional units, thereby capturing the multi-scale characteristics of visual semantic information processing more effectively.

In the first layer, given an input feature vector $\mathbf{x} \in \mathbb{R}^{4096}$, each feature element is associated with a functional-network label according to the Yeo-7 parcellation. The router computes feature-wise routing logits through a linear transformation:
\begin{equation}
\mathbf{z}_i = W_r \mathbf{x}_i,
\end{equation}
where $\mathbf{z}_i \in \mathbb{R}^{7}$ denotes the routing logits for feature $i$, and $W_r$ are learnable parameters shared across features. Stacking the logits over all 4096 features yields a routing matrix $\mathbf{Z} \in \mathbb{R}^{4096 \times 7}$. Applying the softmax function along the expert dimension yields the raw routing probabilities:
\begin{equation}
\mathbf{P}_{\text{raw}} = \text{Softmax}(\mathbf{Z}).
\end{equation}

Based on the Yeo-7 brain atlas, we construct a deterministic prior distribution $p_{\text{roi}}$ for each feature. If the feature is associated with brain region $\text{roi}(i) \in \{1,\ldots,7\}$, the prior distribution is defined as:
\begin{equation}
p_{\text{roi}}^{(i)}[k] = \begin{cases}
1.0, & \text{if } \text{roi}(i) = k, \\
0.0, & \text{otherwise}.
\end{cases}
\end{equation}
This produces a sparse one-hot prior that encodes the correspondence between each feature and its associated brain region expert, such as features derived from voxels in the visual cortex being guided toward the visual region expert.

To regulate the routing distribution toward the network parcellation prior, we incorporate a time-dependent KL divergence regularizer\cite{ref_Yu2013}:
\begin{equation}
\mathcal{L}_{\text{kl}} = w_{\text{kl}}(t) \cdot D_{\text{KL}}\left(p_{\text{roi}}^{(i)} \| \mathbf{P}_{\text{raw}}^{(i)}\right),
\end{equation}
where $w_{\text{kl}}(t)$ varies dynamically with training epoch $t$. During the early training stages, this weight is gradually increased, allowing the model to prioritize learning task-specific features without being overly constrained initially. In the intermediate phase, the weight is maximized to incorporate prior knowledge and guide the model toward routing features in accordance with brain-region assignments. Toward the late stages, the weight is decreased to afford the model greater flexibility for refinement. This staged strategy balances prior guidance with task-driven learning. The KL term is computed feature-wise and aggregated across all features.

To prevent load imbalance and reduce computational redundancy in the expert modules, we adopt a capacity-limited expert-wise Top-K feature selection strategy\cite{ref_Nguyen2023}. For each expert, only the top-$K$ features with the highest routing scores are selected, where
\begin{equation}
K = \left\lceil \frac{L \times CF}{E} \right\rceil,
\end{equation}
with $L=4096$ denoting the total feature dimension, $E=7$ the number of first-layer experts, and $CF \in (0,2]$ the capacity factor. The capacity is computed with respect to the full feature dimension to ensure that each expert processes a balanced subset of informative features under a fixed input size. This routing strategy limits redundant computation while preserving sufficient diversity in the selected feature subsets for expert-specific representation learning.

For expert $k$, the routed input is obtained by selecting and weighting the corresponding feature subset:
\begin{equation}
\mathbf{Q}_k = \mathbf{w}_k \odot (\mathbf{m}_k \odot \mathbf{x}),
\end{equation}
where $\mathbf{m}_k$ denotes the top-$K$ binary mask for expert $k$, and $\mathbf{w}_k$ represents the corresponding routing weights. The weighted feature $\mathbf{Q}_k$ is then passed through the expert's first-layer network:
\begin{equation}
\mathbf{F}_k^{(1)} = E_k^{(1)}(\mathbf{Q}_k).
\end{equation}

In the second layer, experts operate without explicit prior constraints and are driven primarily by data to model fine-grained features. The second-layer expert number is set to 14 to provide sufficient capacity for fine-grained sub-network composition, enabling the model to capture cross-network interactions beyond the coarse anatomical partition. This design facilitates the adaptive capture of high-level semantic features, including inter-network sharing and complex feature combinations, thereby reducing over-reliance on anatomical priors and supporting richer semantic representations.

\subsubsection{Feature Fusion and Mapping}
In each layer, the outputs from all expert modules are aggregated via summation to form local fusion features. This process facilitates cross-scale integration, enhancing the model's capacity to capture fine-grained semantic information. Subsequently, a global pooling layer is applied to the local fusion features to extract global contextual information across regions. Building upon this, the model introduces a diffusion prior mechanism\cite{ref_Fei2023,ref_Wan2024} to perform spatial smoothing on the fused features, aiming to obtain more semantically consistent predictive features. Finally, these predicted features are aligned with the corresponding true CLIP text or image features, effectively enabling multimodal information fusion and matching, which provides a robust semantic foundation for subsequent tasks.

\subsection{Fine-tuning with Spatio-Temporal Routing}
Building upon fMRI feature mapping, we adopt the MORE-Brain framework \cite{ref_Wei2025}, leveraging pre-trained brain encoders and diffusion priors to fine-tune SDXL for brain-to-image generation. At its core is a spatiotemporal routing mechanism that adaptively fuses multi-scale brain representations. Conditioned on the UNet timestep embeddings, a temporal router dynamically modulates feature granularity: aligning with the coarse-to-fine nature of the diffusion process, it prioritizes coarse-grained experts for global semantics in early denoising steps, shifting to fine-grained experts for intricate details in later iterations. Subsequently, a spatial router refines this fusion through cross-attention, allowing the noisy latent $z_t$ to spatially query the temporally gated brain features. This joint routing strategy ensures coherent multi-scale integration and precise spatiotemporal alignment throughout the generation process.
\subsection{Loss Functions}
Our framework is trained using a multi-objective optimization strategy. In addition to the KL divergence constraint employed for brain network expert routing, we integrate several objectives: geometric alignment, SoftCLIP-based contrastive learning, and diffusion priors.
\subsubsection{Geometric Alignment}
To ensure both directional consistency and magnitude scaling of the features, we simultaneously constrain the brain features $b$ and visual semantic targets $c$ using cosine similarity\cite{ref_Minnema2019} and Mean Squared Error (MSE)\cite{ref_Zho2025}:

\begin{equation}
\mathcal{L}_{\text cos}(b, c) = 1 - \frac{b \cdot c}{|b| |c|}, \quad \mathcal{L}_{\text MSE}(b, c) = \frac{1}{N} \sum_{i=1}^{N} (b_i - c_i)^2
\end{equation}

\subsubsection{SoftCLIP Contrastive Learning}
To achieve flexible alignment between brain representations and visual semantic targets, we employ a bidirectional SoftCLIP loss\cite{ref_Scotti2023}:
\begin{equation}
\mathcal{L}_{\text SoftCLIP}(b, c) = - \sum_{i=1}^N \sum_{j=1}^N \left[ \frac{\exp (c_i \cdot c_j/\tau)}{\sum_{m=1}^N \exp (c_i \cdot c_m / \tau)} \cdot \log \left( \frac{\exp (b_i \cdot c_j/\tau)}{\sum_{m=1}^N \exp (b_i \cdot c_m / \tau)} \right) \right]
\end{equation}

\subsubsection{Diffusion Prior and Refinement}
We further introduce a diffusion prior along with a prior contrastive refinement loss\cite{ref_ZEB2025}. This provides robust semantic constraints on the brain features while ensuring high-fidelity generative output. The diffusion loss is defined as:
\begin{equation}
\mathcal{L}_{\text DP} = \mathbb{E}{x_0, \epsilon, t} \left[ |\epsilon - \epsilon_\theta(x_t, t, b)|^2 \right]
\end{equation}
To further refine the predicted output $\hat{c}$ of the prior network, a prior contrastive loss is applied:
\begin{equation}
\mathcal{L}_{CLIP}^{prior} = \text{SoftCLIP}(\hat{c}, c) \times \lambda{prior}
\end{equation}
The total training objective is defined as:
\begin{equation}
\mathcal{L}_{\text total} = \mathcal{L}_{\text kl} + \mathcal{L}_{\text cos} + \mathcal{L}_{\text MSE} + \mathcal{L}_{\text SoftCLIP} + \mathcal{L}_{\text DP} + \mathcal{L}_{CLIP}^{prior}
\end{equation}
\subsection{Interpretability}
To bridge the gap between neural decoding and cognitive neuroscience, we leverage the latent feature space of the pre-trained CLIP model to align decoded brain features with visual semantics. Specifically, we calculate the cosine similarity between the brain features of each expert and the image patch features extracted by CLIP, mapping these similarities into spatial heatmaps to visualize the image regions prioritized by different experts. For the $k$-th expert, the similarity is computed as:

\begin{equation}
S_k = \frac{\mathbf{F}_k \cdot \mathbf{F}_{\text{CLIP}}}{\|\mathbf{F}_k\| \|\mathbf{F}_{\text{CLIP}}\|}
\end{equation}

where $\mathbf{F}_k$ denotes the brain features of the $k$-th expert, and $\mathbf{F}_{\text{CLIP}}$ represents the image patch features extracted by CLIP.

Furthermore, we employ the routing weights of the MoE layer to quantify the relative contribution of each brain functional network to the final decision-making process.
\section{Experiments}
\subsection{Experimental Setup}
We evaluated our model on the NSD dataset using whole-brain 3D fMRI data. 
Data underwent standard preprocessing and registration to align individual 
brain anatomies to a common template space.To ensure a fair comparison, we adopted the same train/validation/test split protocol as MindBridge.
\begin{figure}[H]
\centering
\includegraphics[width=\textwidth]{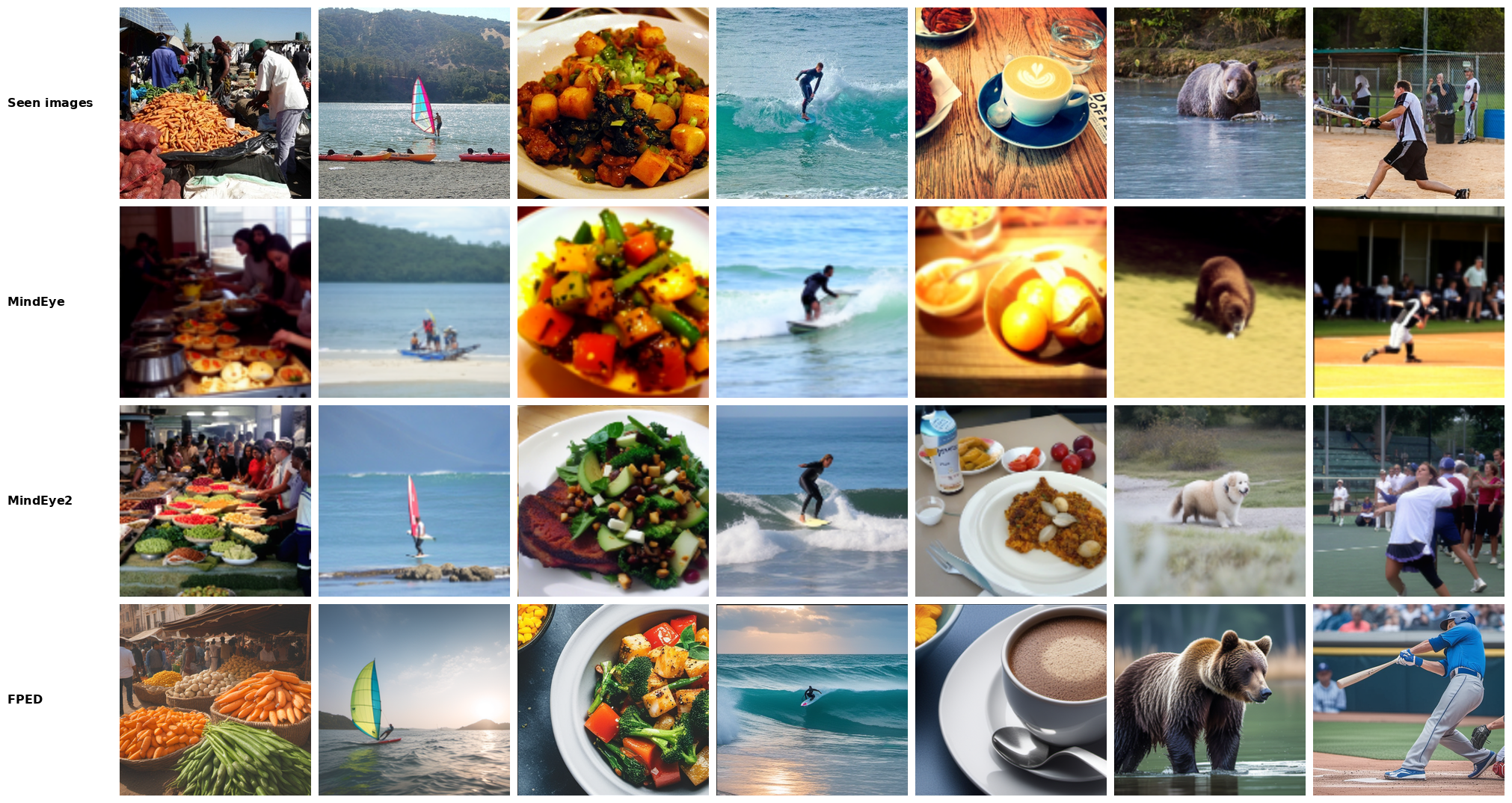}
\caption{Brain-to-image reconstruction performance.} 
\label{fig:recon}
\end{figure}
We employed multiple metrics to comprehensively evaluate reconstruction 
quality. Low-level features were assessed using PixCorr, SSIM, and 
AlexNet (L5), while high-level semantic features were evaluated using 
Inception, CLIP, EffNet-B, and SwAV \cite{ref_MindBridge2024}.

We compared against state-of-the-art baselines: Brain-Diffuser~\cite{ref_Chen2023}, 
MindBridge, and MindEye2 for brain decoding, and MindEye/MindEye2 
for image reconstruction. All reported quantitative metrics and reconstructed images for these baseline methods were taken from their respective original publications.

\subsection{Reconstruction Results}

The image reconstruction results are presented in Figure~\ref{fig:recon}. 
Visually, our framework exhibits remarkable proficiency in capturing 
high-level semantic content while accurately preserving object relationships 
within complex visual scenes.

\begin{table}[H]
\centering
\caption{Comparison of different methods on brain activity decoding tasks.}
\label{tab:comparison}
\begin{tabular}{l|ccc|cccc|c}
\hline
{Method} & \multicolumn{3}{c|}{Low-Level} & 
\multicolumn{4}{c|}{High-Level} & {Params$\downarrow$} \\
\cline{2-8}
 & PixCorr$\uparrow$ & SSIM$\uparrow$ & AlexNet(5)$\uparrow$ & 
Incept$\uparrow$ & CLIP$\uparrow$ & EffNet-B$\downarrow$ & SwAV$\downarrow$ & \\
\hline
Brain-Diffuser & \underline{0.254} & 0.356 & \underline{96.2}\% & 
87.2\% & 91.5\% & .775 & .423 & — \\
MindBridge & 0.151 & 0.263 & 95.5\% & 92.4\% & \underline{94.7}\% & 
.712 & \underline{.418} & \underline{0.69B} \\
MindEye2 & \textbf{0.322} & \textbf{0.431} & \textbf{98.6\%} & 
\textbf{95.4\%} & 93.0\% & \textbf{.619} & \textbf{.344} & 2.58B \\
Ours (single) & 0.109 & 0.381 & 87.7\% & 94.0\% & 96.1\% & .686 & 
.455 & \textbf{0.68B} \\
Ours (multi) & 0.240 & \underline{0.386} & 85.0\% & \underline{94.2\%} 
& \textbf{96.7\%} & \underline{.663} & .421 & \textbf{0.68B} \\
\hline
\end{tabular}
\end{table}

Quantitatively, as summarized in Table~\ref{tab:comparison}, our 
approach achieves state-of-the-art on high-level semantic metrics: 
CLIP score of \textbf{96.7\%}, Inception score of \textbf{94.2\%}, 
and EffNet-B distance of \textbf{0.663}. Notably, despite employing 
lightweight parameterization (\textbf{0.68B} parameters) without VAE-based 
generative modeling, our method achieves the second-best SSIM score 
(\textbf{0.386}), demonstrating strong structural consistency while 
maintaining a favorable trade-off between semantic accuracy and 
pixel-level fidelity. This remarkable efficiency validates that 
parameter-efficient architecture with neurobiologically-grounded 
design can rival or exceed substantially larger models.

\subsection{Ablation Study}
We conducted ablation experiments to evaluate each component's 
contribution. We first examined whole-brain necessity by restricting 
reconstruction to visual cortex features (OnlyV baseline). We then 
compared our MoE architecture against alternative aggregation schemes: 
Transformer-based, Attention-based, and Uniform fusion.
As shown in Table~\ref{tab:ablation}, our full model consistently 
outperforms the visual-only baseline across all metrics, demonstrating 
that non-visual brain regions provide essential complementary 
information for visual reconstruction. The MoE architecture also 
surpasses all alternative variants, validating its effectiveness in 
modeling brain dynamics. These results underscore that explicitly 
capturing functional specialization and inter-network cooperation is 
fundamental to brain-to-image decoding.
\begin{table}
\centering
\caption{Ablation study of different architectures on brain activity decoding.}
\label{tab:ablation}
\begin{tabular}{l|ccc|cccc}
\hline
{Method} & \multicolumn{3}{c|}{Low-Level} & 
\multicolumn{4}{c}{High-Level} \\
\cline{2-8}
 & PixCorr$\uparrow$ & SSIM$\uparrow$ & AlexNet(5)$\uparrow$ & 
Incept$\uparrow$ & CLIP$\uparrow$ & EffNet-B$\downarrow$ & SwAV$\downarrow$  \\
\hline
OnlyV & 0.109 & 0.356 & 84.0\% & 89.1\% & 92.8\% & .687 & .470 \\
Transformer & 0.084 & 0.374 & 82.0\% & 93.5\% & 95.0\% & .721 & .522 \\
Attention & 0.045 & 0.354 & \textbf{86.0\%} & 90.2\% & 92.9\% & .714 & .531 \\
Uniform & 0.056 & 0.370 & 81.4\% & 93.2\% & 93.2\% & .715 & .508 \\
Ours (MoE) & \textbf{0.240} & \textbf{0.386} & 85.0\% & 
\textbf{94.2\%} & \textbf{96.7\%} & \textbf{.663} & \textbf{.421} \\
\hline
\end{tabular}
\end{table}
\subsection{Interpretability Analysis}
To demonstrate the biological plausibility and interpretability of our 
MoE framework, we analyze expert specialization in visual feature 
encoding and how the router dynamically allocates tasks across experts.
\begin{figure}[H]
\centering
\includegraphics[width=\textwidth]{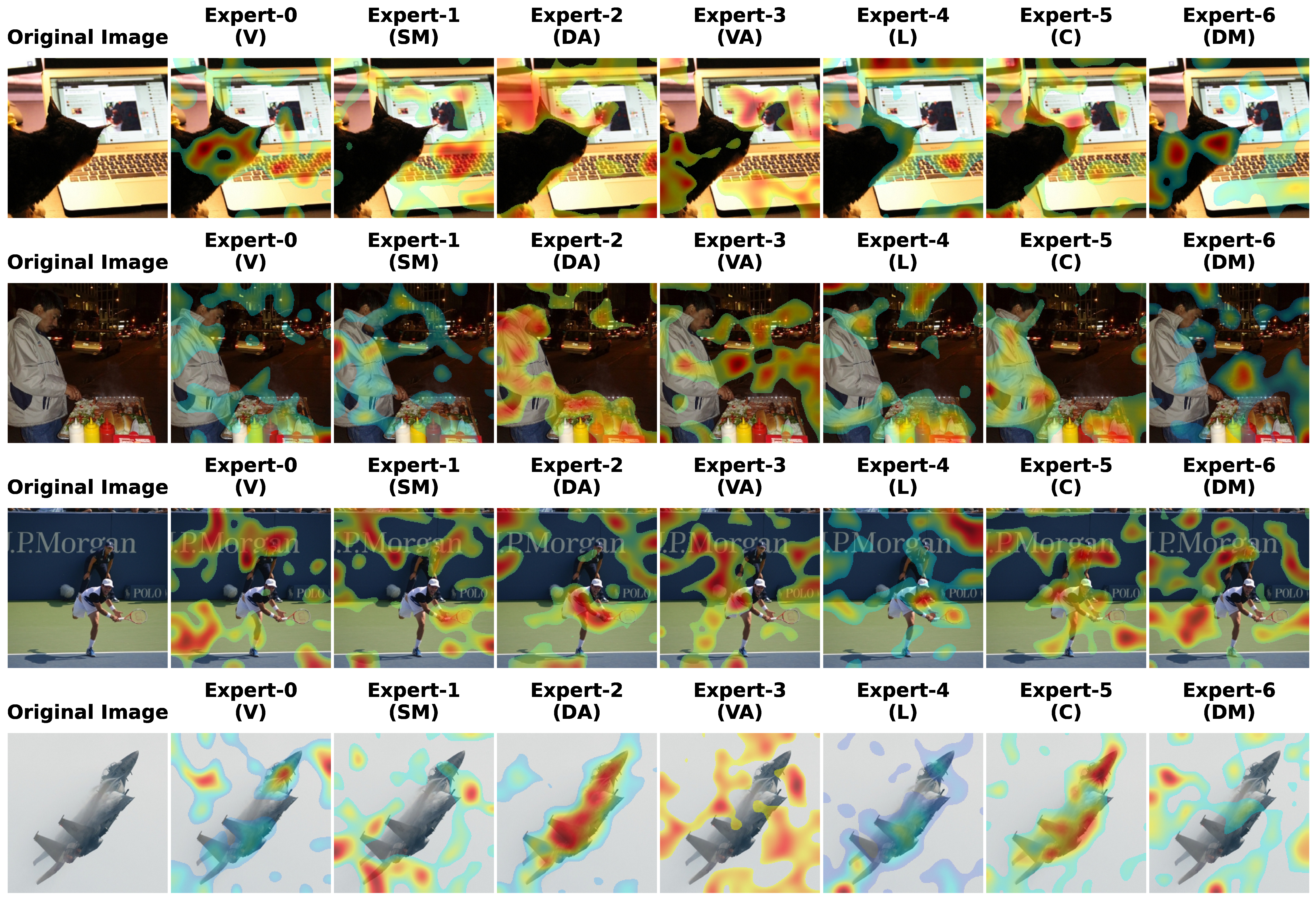}
\caption{Semantic heatmaps of brain-network-based experts. Each row displays 
the original image alongside heatmaps from seven experts (V, SM, DA, VA, L, C, DM).} 
\label{fig:heatmap}
\end{figure}
\subsubsection{Spatial Semantic Specialization}
Visualization of semantic focus (Fig.~\ref{fig:heatmap}) reveals clear 
functional alignment with established brain network organization. V experts exhibit holistic activation across global contours
\cite{ref_KandelER2000}, capturing fundamental structural layouts. In 
contrast, SM and C experts selectively 
activate action-oriented and task-relevant regions
\cite{ref_DosenbachNU2007,ref_HlustikP2001}---such as limbs, tools, and 
operational areas---reflecting their role in motor planning and executive 
control.

Attention-related experts demonstrate functional hierarchy: DA exhibits high-precision focalization on behaviorally salient targets (e.g., athletes or aircraft), while VA displays distributed activation for stimulus-driven reorienting
\cite{ref_CorbettaM2002}. L experts show intense selectivity 
for reward-relevant stimuli (food, faces), whereas DM experts encode integrated scene semantics and contextual information
\cite{ref_MargulieDG2016,ref_PhelpsEA2005}.
Collectively, these patterns reveal consistent functional 
specialization mirroring the modular organization of large-scale brain 
networks---a property difficult to achieve with conventional homogeneous 
decoders lacking network-aware inductive biases. Our framework, by 
enabling conditional expert specialization through dynamic routing, 
yields both superior reconstruction fidelity and biologically 
meaningful brain--semantic correspondence.
\subsubsection{Modality-Specific Routing Behavior}
\begin{figure}[H]
\centering
\includegraphics[width=\textwidth]{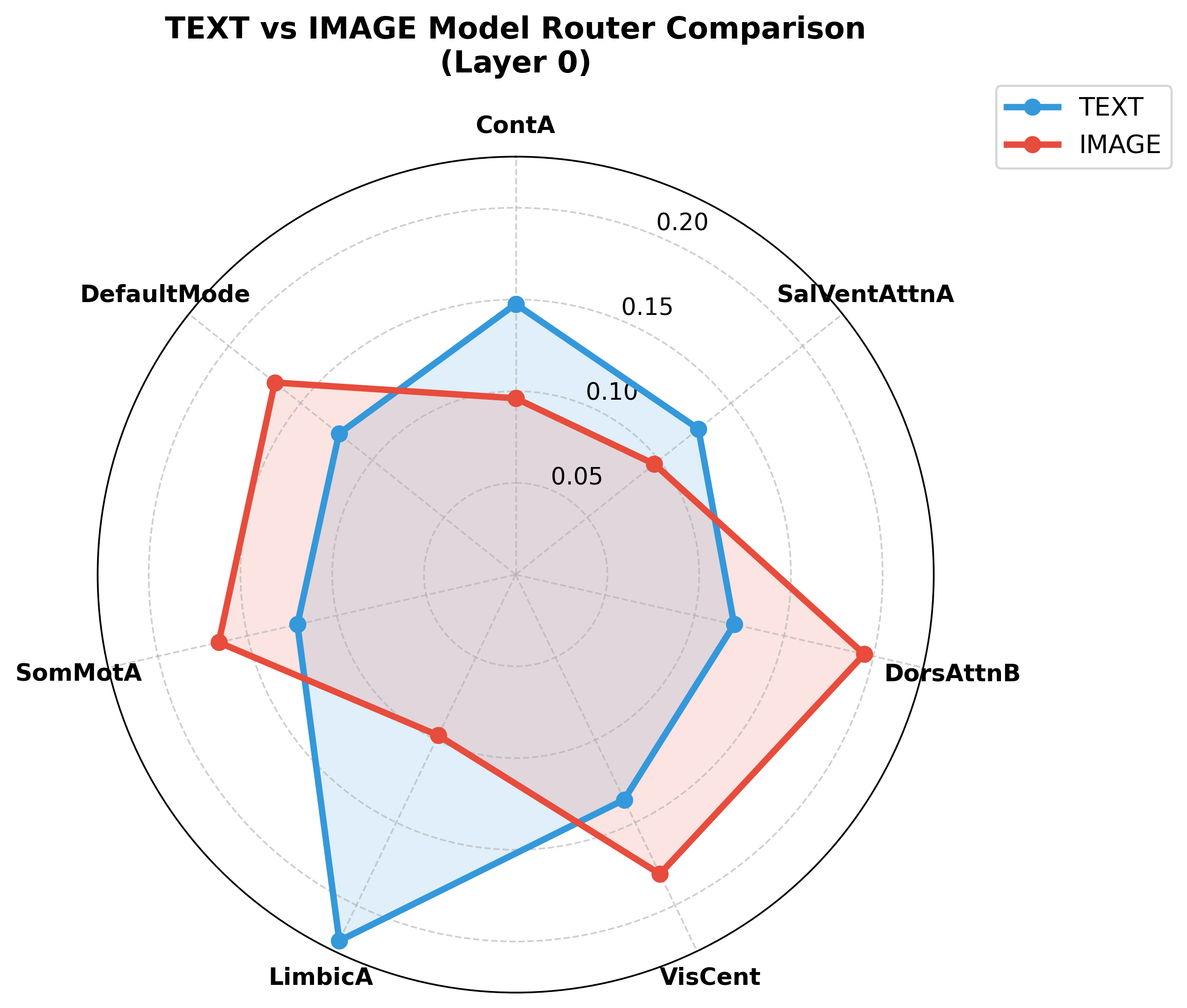}
\caption{Comparison of MoE routing weights across modalities.} 
\label{fig:router}
\end{figure}
The quantification of routing weights (Fig.~\ref{fig:router}) reveals 
distinct neural recruitment strategies for aligning brain features with 
TEXT and IMAGE modalities.
\paragraph{TEXT Alignment:} The L network is the most prominent 
contributor. As a hub for affective and high-level semantic processing, 
its dominance suggests that the model prioritizes abstract semantic 
meaning and emotional connotations when aligning brain representations 
with textual information.
\paragraph{IMAGE Alignment:} The model exhibits multi-network engagement 
involving DA, V, DM, and SM. This balanced recruitment reflects the 
necessity of integrating spatial attention, structural contours, scene 
context, and motor-related semantics when aligning brain features with 
visual information.

Overall, these modality-dependent routing patterns demonstrate that the 
proposed MoE framework adaptively assigns brain networks according to 
functional relevance, providing direct evidence that conditional routing 
aligns with neurobiologically grounded information processing principles.
\section{Conclusion}
In this work, we propose FPED, a novel brain decoding framework that 
integrates ROI priors into a MoE architecture. 
By transitioning from traditional 1D-flattening of visual cortex signals 
to structured whole-brain functional network analysis, our approach 
bridges the gap between neuroscientific interpretability and decoding 
performance. Leveraging the Yeo-7 brain atlas, FPED constrains expert 
modules to specialize in distinct functional networks, thereby mimicking 
the brain's hierarchical processing---from coarse-grained functional 
specialization to fine-grained information integration.

Experimental results demonstrate robust performance across low-level and 
high-level metrics. The dual-layer routing mechanism not only extracts 
complementary features from non-visual cognitive networks (DM and DA) but also provides interpretable 
expert specialization. 

Overall, this work validates that incorporating 
neurobiologically grounded priors into deep learning architectures 
significantly advances brain-to-image reconstruction.
\subsubsection{Acknowledgements} This work was supported by the National Natural Science Foundation of China (Grant. No. 62576275 and Grant.
No.62506298).
%
%
%
%

\end{document}